\documentclass{article}

\PassOptionsToPackage{numbers, compress}{natbib}

\usepackage[final]{neurips_2022}

\usepackage[utf8]{inputenc} 
\usepackage[T1]{fontenc}    
\usepackage{hyperref}       
\usepackage{url}            
\usepackage{booktabs}       
\usepackage{amsfonts}       
\usepackage{nicefrac}       
\usepackage{xcolor}         
\usepackage{graphicx}
\usepackage{caption}
\usepackage{subcaption}
\usepackage{amsmath}
\usepackage{hyperref}

\title{Much Easier Said Than Done: Falsifying the Causal Relevance of Linear Decoding Methods}

\author{%
  Lucas Hayne\textsuperscript{1}\thanks{lucas.hayne@colorado.edu}\\
  \And
  Abhijit Suresh\textsuperscript{1}
  \And
  Hunar Jain\textsuperscript{1}
  \And
  Rahul Kumar\textsuperscript{1}
  \And
  R. McKell Carter\textsuperscript{2}
  \AND
  \textnormal{\textsuperscript{1}Department of Computer Science}\\
  \textsuperscript{2}Department of Psychology and Neuroscience\\
  University of Colorado Boulder\\
}

\begin{document}
\maketitle
\begin{abstract}
  Linear classifier probes are frequently utilized to better understand how neural networks function. 
  Researchers have approached the problem of determining unit importance in neural networks by probing their learned, internal  representations. Linear classifier probes identify highly selective units as the most important for network function. Whether or not a network actually relies on high selectivity units can be tested by removing them from the network using ablation. 
  Surprisingly, when highly selective units are ablated they only produce small performance deficits, and even then only in some cases. In spite of the absence of ablation effects for selective neurons, linear decoding methods can be effectively used to interpret network function, leaving their effectiveness a mystery. 
  To falsify the exclusive role of selectivity in network function and resolve this contradiction, we systematically ablate groups of units in subregions of activation space. 
  Here, we find a weak relationship between neurons identified by probes and those identified by ablation. More specifically, we find that an interaction between selectivity and the average activity of the unit better predicts ablation performance deficits for groups of units in AlexNet, VGG16, MobileNetV2, and ResNet101.  
  Linear decoders are likely somewhat effective because they overlap with those units that are causally important for network function. Interpretability methods could be improved by focusing on causally important units. 
\end{abstract}

\section{Introduction}
Neural networks play critical roles in everything from driving to systems of justice, but trusting neural networks in such roles requires understanding how they function. Widespread adoption of neural networks has raised concerns about trust and bias \cite{Simonite_2018}. For example in medicine, the incorporation of neural networks requires considering how uncertainty in how a decision was made could be argued to undermine medical authority \cite{Grote_Berens_2020}. The requirement for transparency is so widely recognized it has begun to be incorporated in regulation \cite{noauthor_2016-gt}. Improving our understanding of neural network function is in both the interest of the public and AI development communities. Unfortunately, most methods used for interpretation have not been strongly linked to neural network function. 


One approach to understanding neural network function is to examine how they represent information internally \cite{ Arpit_Jastrzebski_Ballas_Krueger_Bengio_Kanwal_Maharaj_Fischer_Courville_Bengio_et_al_2017,bau2017network,morcos2018importance, Raghu_Unterthiner_Kornblith_2021,yosinski2014transferable}.  Internal representations have been shown to be  remarkably useful in interpreting neural networks \cite{kornblith2019similarity,li2015convergent, Bansal_Nakkiran_Barak_2021}. Linear classifier probes \cite{alain2016understanding} are commonly used as a tool for interpreting \cite{bau2017network,kim2018interpretability,adi2016fine,hupkes2018visualisation} and testing internal representations \cite{zhang2016colorful, zhang2017split, tian2019contrastive, anand2019unsupervised}.
Although linear probes have utility for understanding network representations, studies removing (ablating) groups of neurons identified by linear probes have inconsistent outcomes. The best way to confirm the causal importance of groups of neurons is to remove them and see if the network stops performing its function \cite{lecun1989optimal}, hereafter called ablation. Performance is not consistently degraded when ablating the units most selective for the target class (i.e. units that activate more for the target class than other images) \cite{meyes2020under, morcos2018importance, zhou2018revisiting}. As an example, Donnelly and Roegiest \cite{donnelly2019interpretability} ablate a neuron highly selective for sentiment and observe negligible deficits in performance. This argument can be extended to both training against class selectivity \cite{leavitt2020selectivity} and for the effect of class selectivity on overall performance \cite{kanda2020deleting, zhou2018revisiting, morcos2018importance}. However, there are examples in which researchers observed a weak correlation between class selectivity and ablation deficits \cite{zhou2018revisiting, liu2018understanding}. The significant issue remains that it is difficult to predict the effects caused by ablating the selective units identified by linear probes.

Most ablation studies of performance remove a single unit at a time or are based on single measures of importance (activation strength, selectivity, decoding weight, information-theoretic measures,  etc.) \cite{morcos2018importance,zhou2018revisiting,dalvi2019one,meyes2020under,tu2016reducing, liu2021group, bau2018gan}, preventing study of the aggregate effects of groups of units. Single unit ablation avoids the combinatorially difficult problem of choosing which units to group, but results in small or no changes in performance. Larger performance deficits can be achieved by targeting aggregate groups of units based on a single metric (e.g., selectivity). However,  utilizing single units or single dimensions precludes a systematic exploration of how function and representation vary across activation space. An approach that accounts for these considerations is warranted. In this work, we focus on four heavily studied convolutional neural networks: AlexNet, VGG16, MobileNetV2, and ResNet101. We efficiently group units by their positions in activation space along two dimensions: average class selectivity and average activation magnitude, to highlight the regions occupied by function and representation. This paper documents these contributions: 

\begin{itemize}
    \item We demonstrate a reliable and fast method of grouping units in activation space. This method improves on past clustering work \cite{hod2021detecting} by using a simpler algorithm which clusters units into groups in activation space.
    \item We show that many groups of units in activation space produce high linear classification probe accuracies with higher accuracies for groups of units that are selective for the target and highly selective for not-the-target.
    \item We identify a region of activation space in which ablation selectively reduces performance. This region partially overlaps with parts of activation space that produce strong linear classification. This partial overlap falsifies the specificity of linear probes for isolating function. 
    \item We show that this region is consistent across four popular convolutional neural networks.
\end{itemize}

\section{Methods}
\label{sec:methods}
For our experiments, we investigated AlexNet \cite{krizhevsky2012imagenet}, VGG16 \cite{simonyan2014very}, MobileNetV2 \cite{sandler2018mobilenetv2}, and ResNet101 \cite{he2016deep} all pre-trained on ImageNet \cite{deng2009imagenet}\footnote{AlexNet pre-trained weights from \url{http://github.com/heuritech/convnets-keras}. Other pre-trained models downloaded from \url{http://keras.io/api/applications}}. We modified these pre-trained networks to include lambda masking layers which allowed us to selectively ablate individual units or unit groups in the network. Ablating a unit sets its outgoing activation to zero, preventing information flow from that unit to the next layer. Our process of relating linear probe accuracies and causal network performance for unit groups consists of two steps: 1) Selection of unit groups, 2) Measuring linear probe accuracies and causal performance through ablation, outlined in the following sections.

\subsection{Selection of unit groups}
This section outlines a cheap algorithm for grouping important units in activation space, this allows for all experiments to be run on a laptop, minimizing environmental impact. This algorithm groups units on a reduced activation space constructed using activation characteristics that were implicated in performance based on previous ablation studies. To find groups of units important for class-specific performance in AlexNet, we measured activations for each unit in the network for a set of 2500 images representing 50 classes (e.g., junco bird, brain coral, etc.) drawn from the ImageNet validation set. We extended our findings to three other networks using a subset of 8 image classes. We averaged the activity for each unit for our class of interest and across a random subset of other classes separately for each layer of the network. We then plotted neurons in this two-dimensional activation space with average activation for our class of interest along one dimension and average activation for other classes across the other dimension. 

\begin{figure}
  \centering
     \begin{subfigure}{0.3\textwidth}
     \centering
    \includegraphics[height=1.4in]{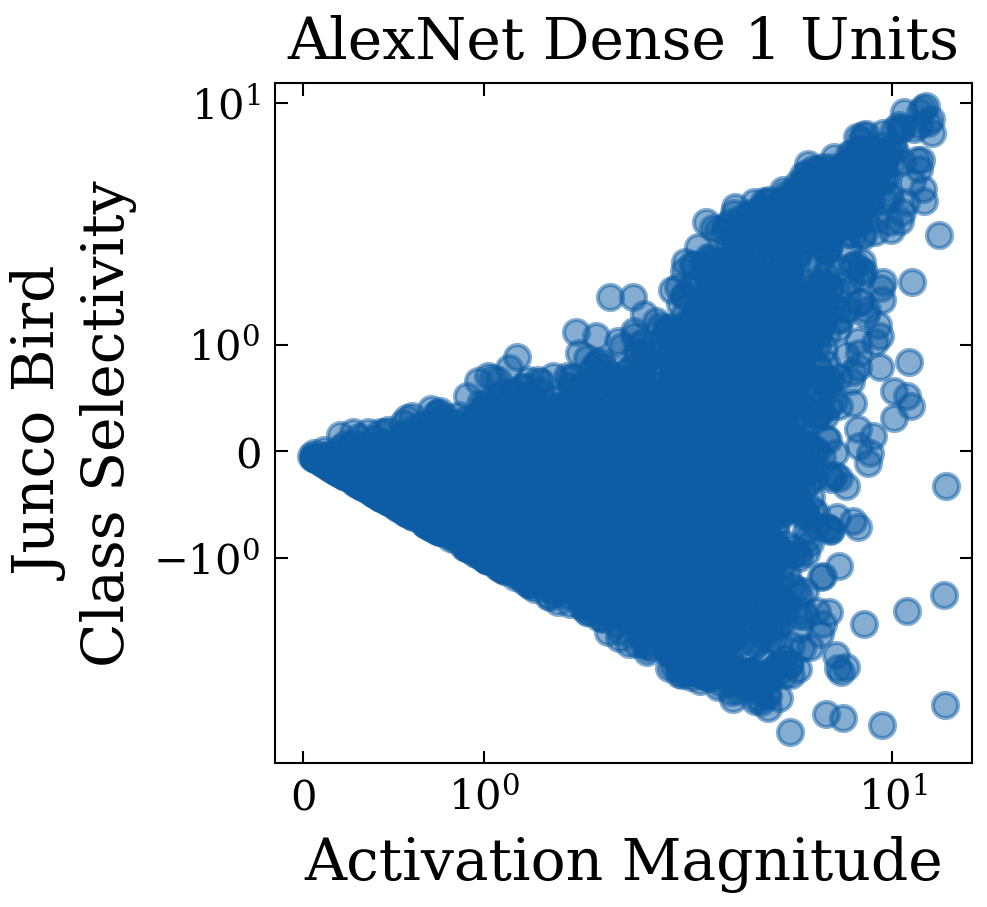}
    \caption{}
    \label{fig:junco_activation_space}
    \end{subfigure}
     \begin{subfigure}{0.3\textwidth}
     \centering
    \includegraphics[height=1.4in]{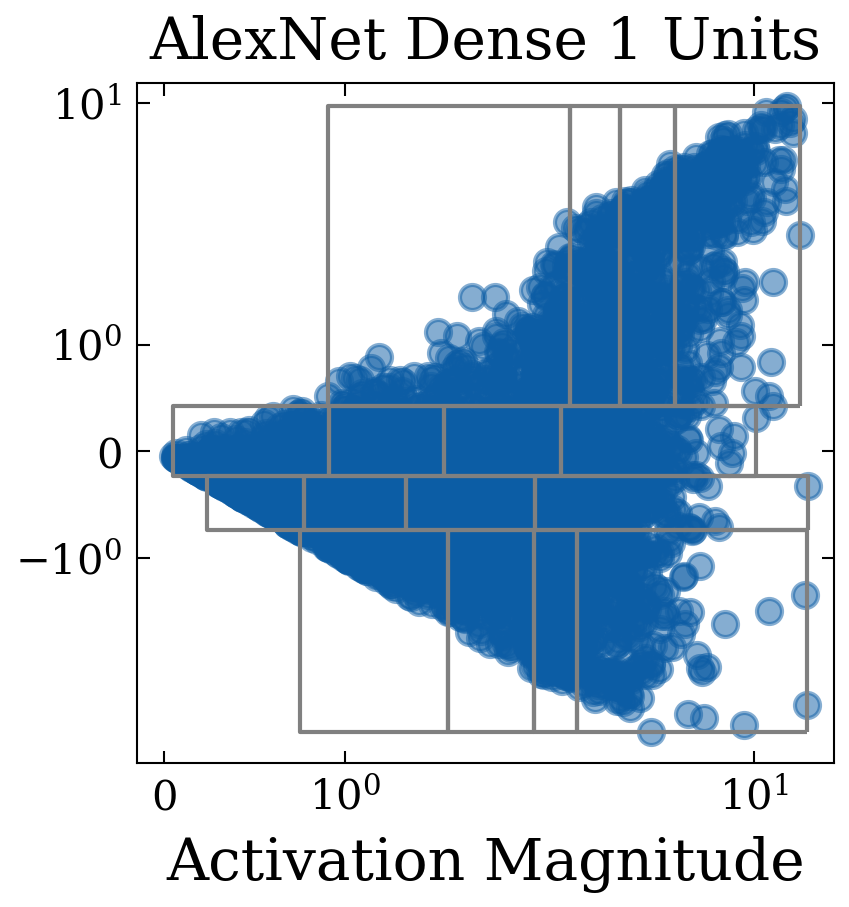}
    \caption{}
    \label{fig:junco_grid}
    \end{subfigure}
     \begin{subfigure}{0.3\textwidth}
     \centering
    \includegraphics[height=1.4in]{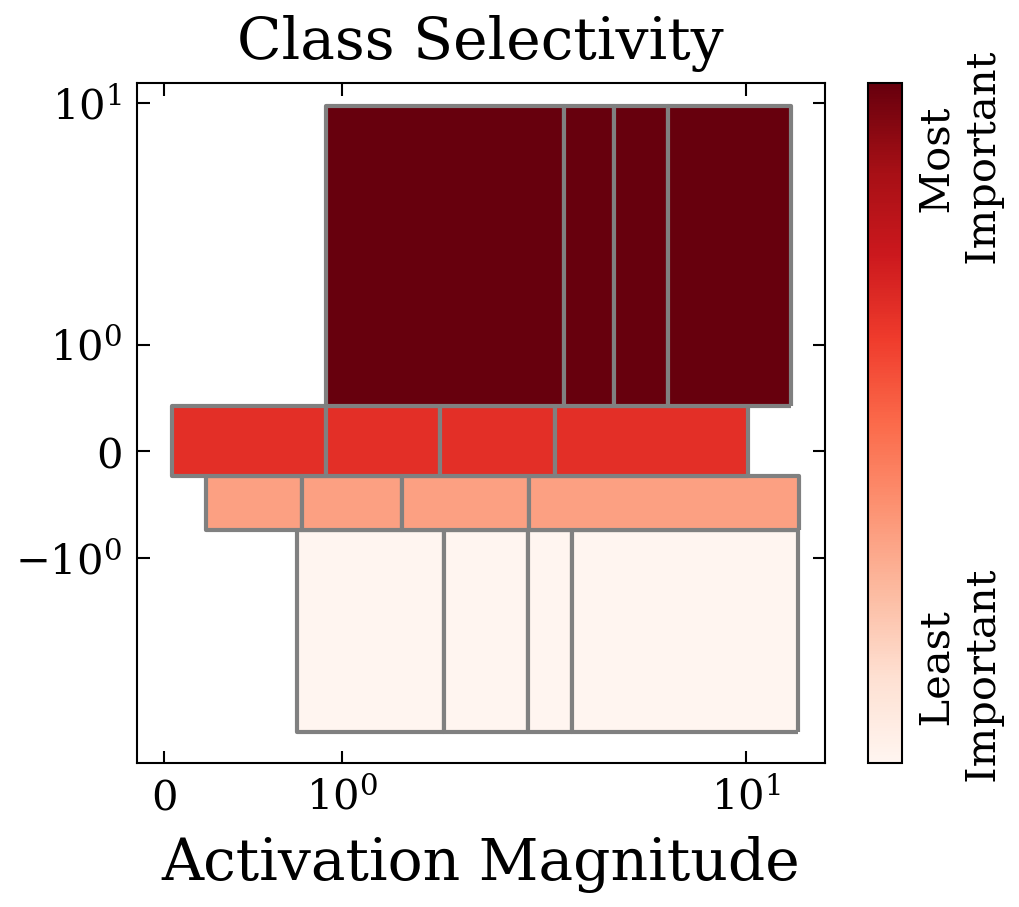}
    \caption{}
    \label{fig:selectivity_hypothesis}
    \end{subfigure}
  \caption{\textbf{Clustering of units by selectivity and magnitude.} \textbf{a)} An example of the reduced activation space for a target class. The vertical axis measures a unit’s average selectivity for the class. The horizontal axis measures a unit’s average magnitude of activation. \textbf{b)} We constructed a grid to overlay on the activation space so that each grid cell contained the same number of units. \textbf{c)} We predict that the most selective unit groups are important for linear probe and causal network performance.}
\end{figure}

We then rotated this space to form two new dimensions: class selectivity and activation magnitude (Figure \ref{fig:junco_activation_space}). In this space, class selectivity measures the difference in the average activation of each unit between our class of interest and other classes. Activation magnitude, on the other hand measures each unit’s average activation in response to all images. Over this space we constructed a 4x4 grid with rectangular cells. The cells vary in area, but each contains an equal number of units. These cells were constructed by first partitioning the space into four horizontal strips at intervals along the selectivity dimension, each with the same number of neurons. These strips were then individually partitioned into four quadrants each along the magnitude dimension so that each of the sixteen resulting cells contained the same number of neurons. Controlling for the number of units in each cell allowed for easy comparisons between cells. Figure \ref{fig:junco_grid} shows the constructed grid overlaid on the activation space. Cells were selected one at a time and all the units from that cell were used to measure linear probing accuracy and causal performance deficits, as outlined in the next subsection.

\subsection{Measuring linear probe accuracies and causal performance deficits}
To measure linear probe decoding accuracies, we followed the methodology presented in the original paper \cite{alain2016understanding}, fitting linear probes to each cell's unit group to predict class identities. For comparison, the importance of each cell’s unit group for class-specific task performance was measured using ablation. First, before ablation, the network’s baseline performance was recorded. During inference on each image, classes were ranked according to the magnitude of their corresponding softmax outputs. For instance, a rank of three for a particular image denotes that the network predicted that the correct class was the third most likely class given the image. After recording baseline ranks, every unit in a given cell was ablated using the lambda mask mentioned previously (see Section \ref{sec:methods}). A performance impact score was calculated for each image by measuring how many ranks the correct class fell compared to baseline, which we refer to as rank deficits. Selective ablation rank deficit scores were calculated by averaging the performance impact scores across the class of interest. A network's top-K classification accuracy depends exclusively on the rank of the correct class, making the rank deficit score a direct measure of the impact of ablation on network performance.

\section{Results}
\begin{figure}
\begin{subfigure}{\textwidth}
     \centering
    \makebox[\textwidth][c]{\includegraphics[width=1.19\textwidth]{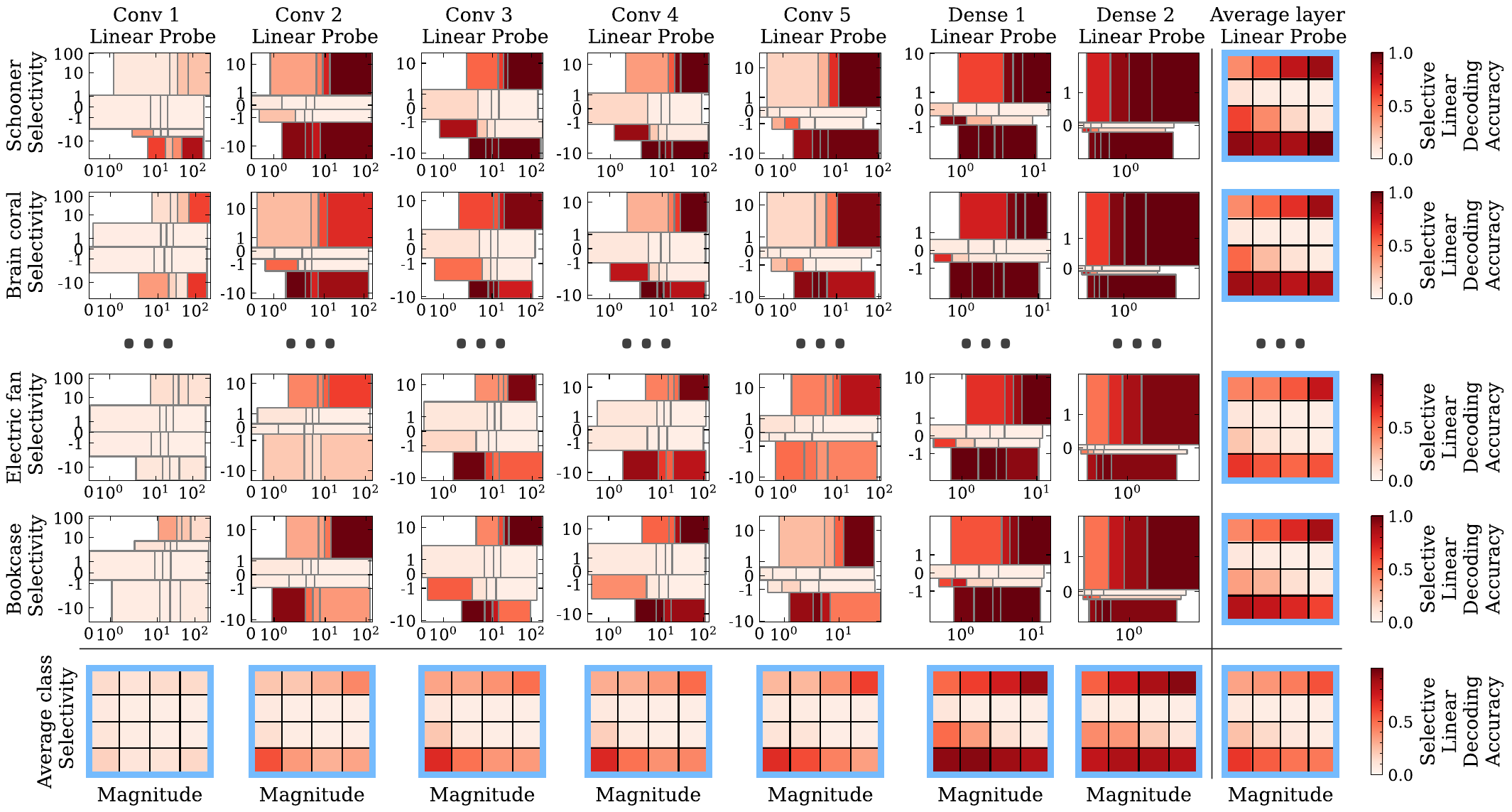}}%
    \caption{Linear Probe}
    \label{fig:srs}
\end{subfigure}
\begin{subfigure}{\textwidth}
    \makebox[\textwidth][c]{\includegraphics[width=1.19\textwidth]{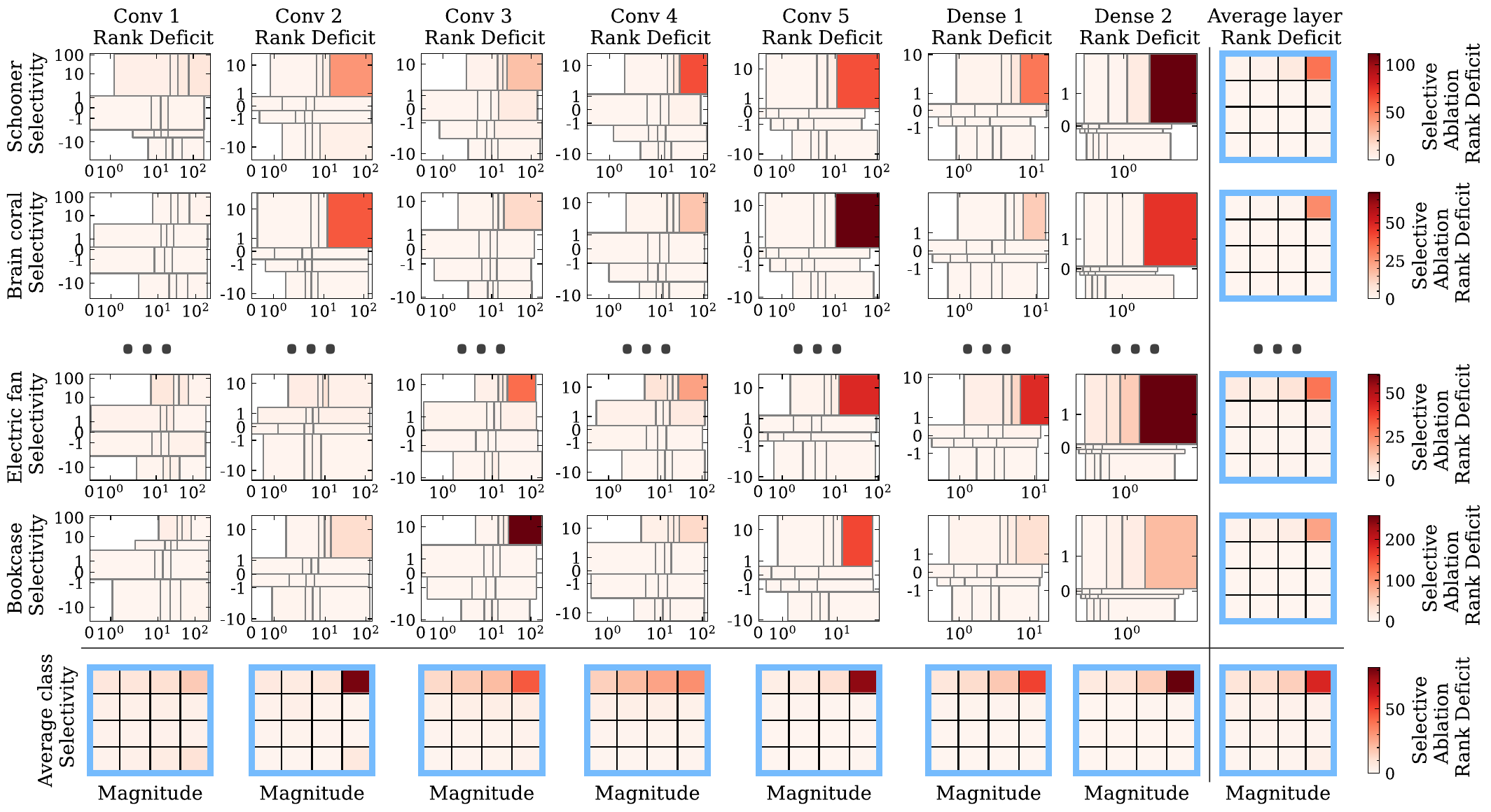}}%
    \caption{Rank Deficit}
    \label{fig:srd}
\end{subfigure}
\caption{\textbf{Linear decoding and ablation scores do not match.} Figure \ref{fig:srs}) Linear probe scores for each class (rows) and each layer of AlexNet (columns). For each class and layer combination we show accuracies obtained from decoding the target class using linear classifier probes on units in each cell. Figure \ref{fig:srd}) Rank deficit scores for each class (rows) and each layer of AlexNet (columns). For each class and layer combination we show class-specific rank deficits caused by ablating all the units in each cell of the grid placed on activation space. Highly selective regions, both for the target class and for all other classes over the target class, consistently produce the highest decoding scores (\ref{fig:srs}), whereas highly impactful neurons tend have high class selectivity and magnitude (\ref{fig:srd}). Average effects of class and layer are shown at the edges to assess consistency of a particular pattern.
}
\end{figure}

\begin{figure}
  \makebox[\textwidth][c]{\includegraphics[width=1.25\textwidth]{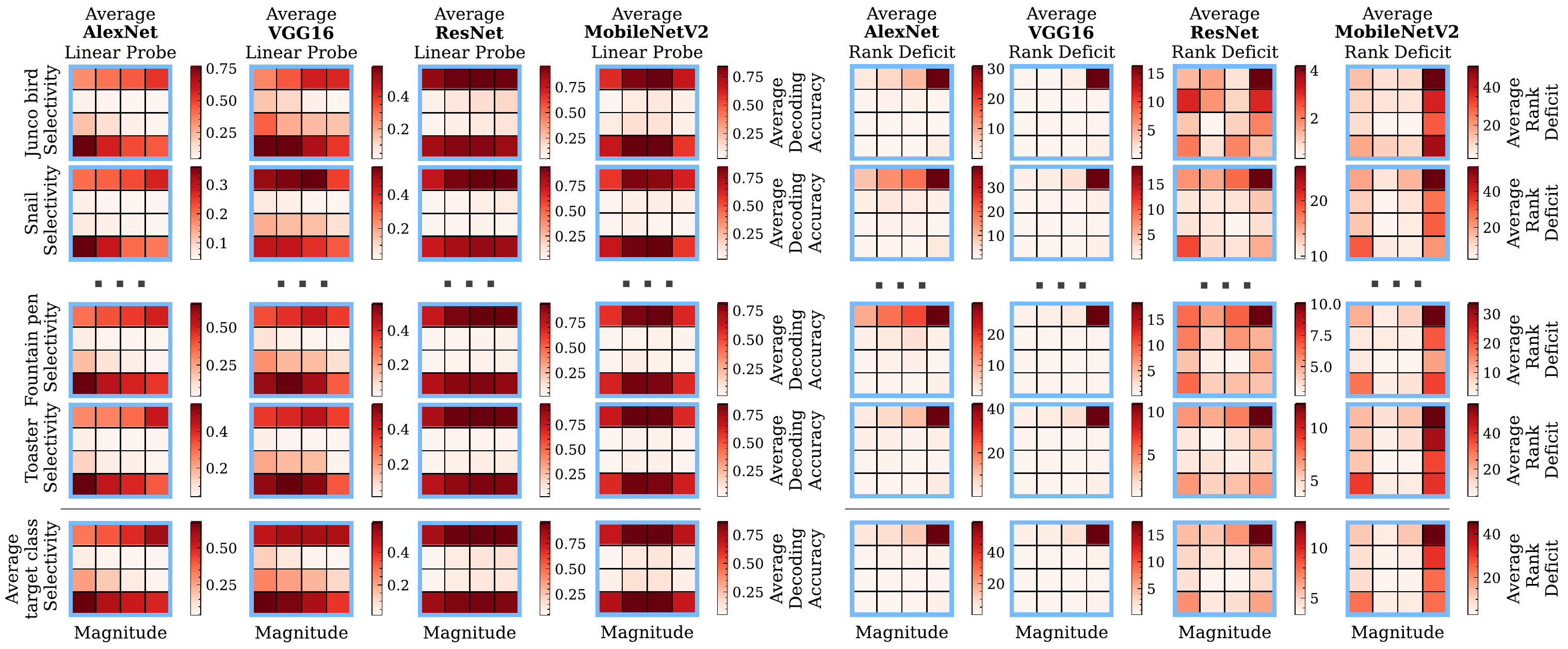}}
  \caption{\textbf{High ablation impact and linear probe accuracy neurons occupy different regions of activation space across four CNNs.} Linear probe accuracies (left) and ablation deficit scores for each network averaged across all layers. For every network, the most selective and active cell produces the highest selective ablation rank deficits. This pattern is most prominent for AlexNet (50 classes) and VGG16 (8 classes), where few other cells contribute to performance. However, in the case of ResNet101 (8 classes) and MobileNet2 (8 classes), other cells can produce high rank deficits. Linear probe accuracies exhibit a distinctly different pattern whereby highly selective cells (both for and against the target class) produce the highest decoding accuracies.}
  \label{fig:network_averages}
\end{figure}

We show grids colored by linear probe accuracies for each image class in our set (rows in Figure \ref{fig:srs}) and for each layer in AlexNet (columns in Figure \ref{fig:srs}). Correspondingly, we show performance deficit grids colored by rank deficit for each image class and each layer (Figure \ref{fig:srd}). The last row depicts the average grid scores for each layer averaged across all 50 classes. The final column depicts the average grid scores for each class averaged across all seven layers of AlexNet. In all cases, the colorbar displays the range in scores for each class. In addition to AlexNet, we show results for three other networks: VGG16, MobileNetV2, and ResNet101. Figure \ref{fig:network_averages} depicts the ablation and linear probe grid scores averaged across all layers for each network we tested.

\subsection{Linear probe accuracy}
Linear probe accuracy is somewhat localized in activation space (Figure \ref{fig:srs}). In general, across all eight image classes, unit groups with maximum accuracy scores are located in the most selective regions of activation space. High linear-probe scores come from cells that are both highly selective for the target class and highly selective on average for all other classes over the target class. Importantly, deleting these neurons has no effect on the networks ability to identify the class, as shown in Figure \ref{fig:srd}. Overall, like ablation impacts, decoding accuracies tend to increase with depth.

\subsection{Rank deficit score}
Rank deficits are localized in activation space. In general, across all image classes, unit groups with a maximum rank deficit are located in the most selective and most active regions of activation space (Figure \ref{fig:srd}). In general the most selective and active unit group produces rank deficits far greater than neighboring unit groups. This pattern shows a combination of selectivity and magnitude contribute to unit importance. This relationship appears discontinuous rather than graded. Even though high impact cells consistently emerge in the most selective and active regions of activation space, the layer with the highest impact cell varies across classes. For instance, the junco bird class has the highest impact cell in the penultimate layer of the network (Dense 2), while the bookcase class shows the highest rank impacts in Conv 3. Additionally, rank deficit scores vary considerably between layers and between classes. Ablating the most selective and most active cell in Conv 3 produces a rank deficit score of nearly 300 for the bookcase class. But, junco class rank deficits never exceed 60.

\subsection{Only a subset of selective unit groups contribute to causal function}
We found the units that are the most active while also being selective are the strongest drivers of performance across all four networks. This pattern is most prominent in AlexNet and VGG16. In ResNet101 and MobileNetV2, the most selective and active cell continues to serve as the strongest driver of performance, but unlike VGG16 and AlexNet, other cells become increasingly important. Particularly, high magnitude cells are consistently important in MobileNetV2. MobileNetV2 and VGG16 both utilize batch normalization layers which have been shown to spread out class-specific information across units \cite{zhou2018revisiting}. Further investigation is needed to establish explicit links between regularizers, like batch normalization, representation, and performance.

\section{Discussion}
In this work, we used a computationally inexpensive method for selecting important groups of units in neural networks. Ablating these groups revealed a consistent pattern whereby the most important unit groups emerge in highly selective and highly active areas of activation space. Previous ablation studies may have failed to identify this pattern for the following reasons. Morcos et al. \cite{morcos2018importance} and Zhou et al. \cite{zhou2018revisiting} ablate single units across the whole network based on class selectivity alone, but not activation magnitude. In Figure \ref{fig:srd}, selectivity alone would have a weak effect on performance because it averages both strong and weak effects, one potential explanation for the weak relationship between selectivity and performance from \cite{zhou2018revisiting}. In contrast, Meyes et al. \cite{meyes2020under} visualize both class selectivity and class-specific magnitudes, but not per layer. Since units from early layers are typically more active and selective, clusters in activation space containing units from all layers lump together highly selective and active units from later layers with less selective and active units from early layers. In this way, network-wide groups are too heterogeneous to reveal patterns that exist primarily at the layer level. This insight is leveraged by Liu et al. \cite{liu2018understanding} to find highly impactful neurons using selectivity (and other promising measures) for small MLPs. Here, we show that only the most selective and active groups of units contribute to causal performance in four large CNNs.

Rather than ablating units, most neural network interpretability studies rely on interpreting function from internal representations. Alain et al. \cite{alain2016understanding} proposed using linear classifier probes to extract the semantic concepts learned by specific units. We use linear probes to show that the most selective units in activation space do indeed contain the most decodable semantic information. However, ablating these highly selective units in activation space does not necessarily produce the largest performance deficits. In fact, only a relatively small subset of selective units produce performance deficits. Across four popular CNNs, high ablation impact neuron groups tend to have both high selectivity and magnitude. This finding suggests that interpretability methods which rely on linear probes may be seeking out directions in activation space with only minimal effects on network performance. We improve on past work by identifying a region of activation space crucial for interpreting function.

\section{Limitations}
\label{sec:limitations}
Limitations in this work stem from three assumptions made in our methods. First, we only investigate class-specific selectivity, while other studies have emphasized that class-specific selectivity can play a different role in network function from low-level feature selectivity \cite{ukita2020causal}. Additionally, other importance measures (i.e. mutual information \cite{liu2018understanding}) are not studied here, but should be the focus of future study. Second, our unit grouping method relies on handpicked hyperparamters that determine the number of cells and produces cells that contain different numbers of units in each layer, making it difficult to compare scores across layers. Lastly, we study the performance changes at the output of the network induced by ablation, while neglecting other causal changes in the rest of the network.

\section{Conclusion}
This work contributes directly to the interpretability of neural networks. We demonstrate a reliable and fast method of grouping units in activation space. We identify a region of activation space in which ablation selectively reduces causal performance, consistent across four popular CNNs. Surprisingly, we found that important ablation impact units only occupy a relatively small subset of the activation space covered by selectively important units identified by linear probes. As a result, interpretability tools that rely on decoding semantic information likely only weakly relate to performance. Future investigations would benefit from extending this research to more networks and tasks to uncover theoretical relationships between neural network representation and function.


\begin{ack}
This work was supported by the University of Colorado Boulder Computer Science Department.

\end{ack}

\small
\bibliography{neurips_2022}

\end{document}